\pdfoutput=1
\documentclass{article}

\usepackage{PRIMEarxiv}

\usepackage[utf8]{inputenc} 
\usepackage[T1]{fontenc}    
\usepackage{hyperref}       
\usepackage{url}            
\usepackage{booktabs}       
\usepackage{amsfonts}       
\usepackage{nicefrac}       
\usepackage{microtype}      
\usepackage{lipsum}
\usepackage{fancyhdr}       
\usepackage{graphicx}       
\graphicspath{{media/}}     

\usepackage{algpseudocode}
\usepackage{algorithm}
\usepackage{color}

\title{Othello is solved
}

\author{
  Hiroki Takizawa \\
  Preferred Networks, Inc. \\
  Chiyoda-ku, Tokyo, Japan\\
  \texttt{contact@hiroki-takizawa.name}
}

\begin{document}
\maketitle

\begin{abstract}
The game of Othello is one of the world's most complex and popular games that has yet to be computationally solved. Othello has roughly ten octodecillion (10 to the 58th power) possible game records and ten octillion (10 to the 28th power) possible game positions. The challenge of solving Othello, determining the outcome of a game with no mistake made by either player, has long been a grand challenge in computer science. This paper announces a significant milestone: Othello is now solved. It is computationally proved that perfect play by both players lead to a draw. Strong Othello software has long been built using heuristically designed search techniques. Solving a game provides a solution that enables the software to play the game perfectly.
\end{abstract}


\begin{figure}[htb]
\centering
\includegraphics[width=1.0\linewidth]{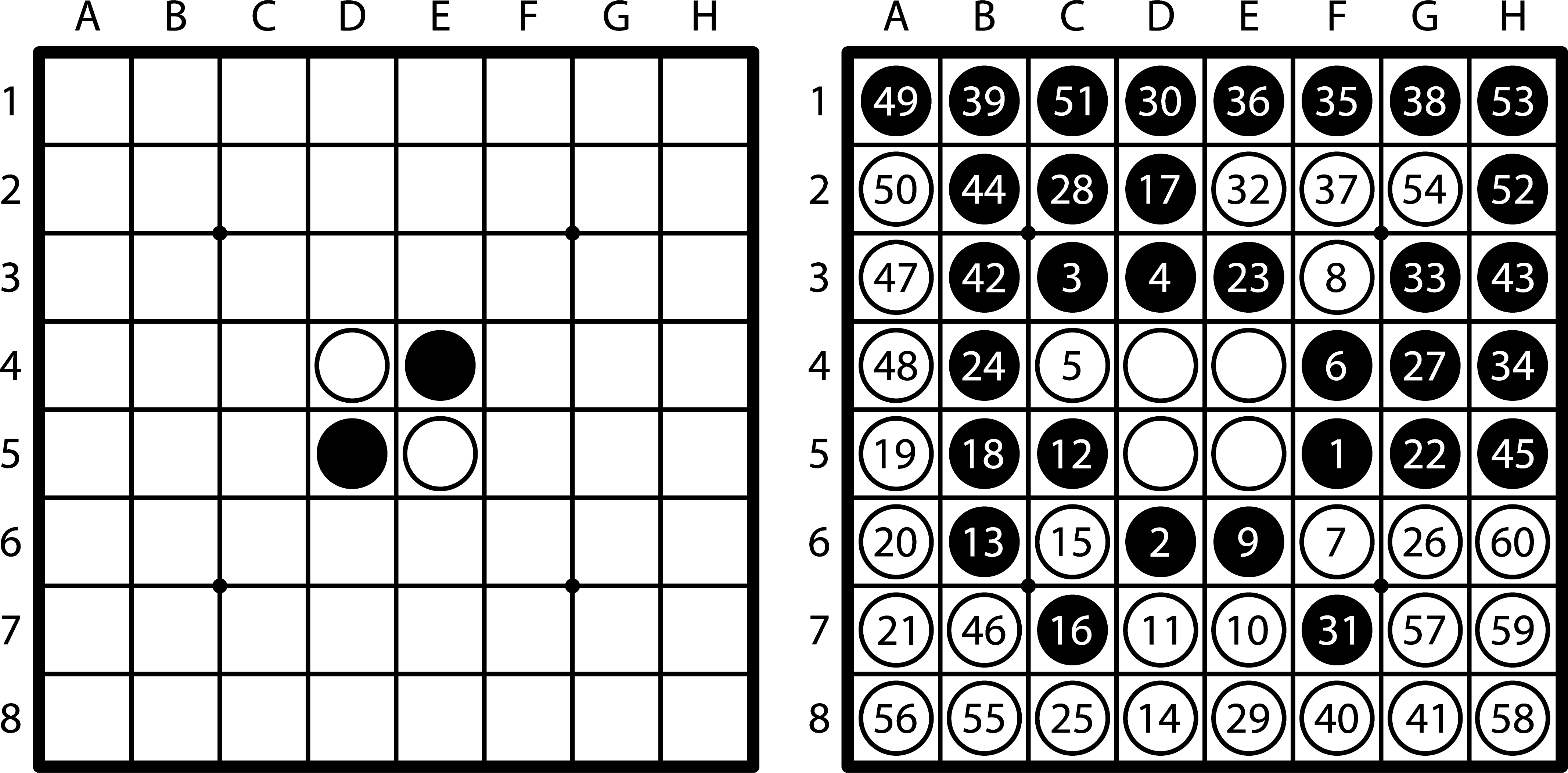}
\caption{
(Left) The initial board position of $8\times 8$ Othello. (Right) A diagram of an optimal game record designated by our study. The game record is ``F5D6C3D3 C4F4F6F3 E6E7D7C5 B6D8C6C7 D2B5A5A6 A7G5E3B4 C8G6G4C2 E8D1F7E2 G3H4F1E1 F2G1B1F8 G8B3H3B2 H5B7A3A4 A1A2C1H2 H1G2B8A8 G7H8H7H6''. The numbers in the stones indicate the order of moves, and the colors of stones indicate the final result. Our study confirms that if a deviation from this record occurs at any point, our software, playing as the opponent, is guaranteed a draw or win. 
}
\label{fig:fig_game_record}
\end{figure}

\newpage

\section{Introduction}
Mastering pure strategy games like chess has been considered a symbol of human intelligence. Since the dawn of computer science, this has been a subject of artificial intelligence (AI) research. For example, there were early consideration by Charles Babbage \cite{babbage1864passages} and Claude Shannon \cite{shannon1950xxii}. To date, with the enhancement of machine learning techniques and computing capabilities, superhuman-strength software has been developed for some of the most popular games, including chess \cite{campbell2002deepblue}, Go \cite{silver2016mastering-alphago}, Shogi (Japanese chess) \cite{kaneko2019shogi, alphazeropaper}, and Othello \cite{buro1997othello}. However, these superhuman-strength programs cannot perfectly solve the games.

Perfectly solving these games (called games of perfect information) means to determine the final result, which is the outcome of the game under perfect play by both players; this result is termed the ``game-theoretic value''. Solved games are classified into at least three types \cite{allis1994searching-thesis, takizawa2023strongly}. The most basic type is called ultra-weakly solved games. In this category, we know the game-theoretic value of the initial board position but not any actual winning strategy. Next, in the case of weakly solved games, we know not only the game-theoretic value of the initial position but also a strategy for both players to achieve this value from the initial position under reasonable computational resources. For example, checkers was weakly solved in this sense \cite{schaeffer2007checkers}. At more comprehensive level, we have strongly solved games, where the outcomes are calculated for all possible position that might arise during game-play.

Othello (also called Reversi) is a highly popular game due to its deep strategic nature. It was invented in the 19th century in England, and in the 20th century, the current format of Othello became widespread in Japan and is now played all over the world. The annual World Championships have been held since 1977, which demonstrates its widespread appeal across the globe. 

One of the reasons for Othello's strategic richness is its vast exploration space. Assuming that there are approximately 10 average moves in each position of the game and an overall average of 58 moves throughout a game, the total number of possible game records can be estimated to be around $10^{58}$, and the total number of possible board positions was also estimated to be around $10^{28}$ \cite{allis1994searching-thesis}. These values are far larger than those of any game that has been solved as a grand challenge up to now, including checkers \cite{schaeffer2007checkers}. As a result, Othello has remained unsolved.

In this paper, we announce that we have weakly solved Othello ($8\times 8$ board). The game-theoretic value of the initial position turned out to be a draw (an optimal game record and the final result are shown in Figure \ref{fig:fig_game_record}). This is not surprising because human Othello experts already predicted it. Another notable point is that the number of positions we needed to explore to get the strict solution was far less that predicted in previous research\cite{allis1994searching-thesis}. We believe this is due to our sophisticated search algorithm configuration.

The Othello result is a monumental achievement for humanity, demonstrating the remarkable advances in computer science and AI technology. Solving Othello has been one of the grand challenges for AI. Over recent decades, AI capabilities have expanded owing to advances in both computing power and algorithms, including enhanced search techniques. In our study, even with the use of the latest computer cluster, solving Othello remained a significant hurdle. Our breakthrough came by improving search efficiency and modifying the latest Othello software.

This paper describes our method to solve Othello, several findings as results, and implications of this research. The raw data and programs to reproduce the results are available on GitHub, Zenodo, and figshare (see Data Availability section).

\section{Related Works}

\subsection{Solved Games}
To the best of our knowledge, the latest game solved prior to this study as a grand challenge is checkers \cite{schaeffer2007checkers}. However, many nontrivial games have been solved, including Connect Four \cite{allis1994searching-thesis}, Qubic
\cite{allis1994searching-thesis}, Go-Moku \cite{allis1994searching-thesis}, Nine Men’s Morris \cite{gasser1996solving-nine-mens-morris}, and 
Awari \cite{romein2003solving-awari}. The difficulty of solving these games largely depends on the number of positions or situations in the game. Solving a game not only reveals its outcome but can also be useful in creating puzzles based on that game \cite{takizawa2023strongly}.

\subsection{Solving Technique}
The algorithms used to solve games have been extensively studied, and they are chosen based on the purpose and nature of the game. For weak solutions, alpha-beta search \cite{knuth1975analysis-alphabeta} is often used, while retrograde analysis \cite{thompson1986retrograde} is frequently used for strong solutions. Additionally, algorithms like depth-first proof-number (df-pn) search \cite{nagai2002df-pn, kishimoto2012game},which is based on proof-number (pn) search \cite{ALLIS1994pnsearch}, have been developed to solve puzzles with very long solution sequences. In our study, we utilized alpha-beta search because our goal was to obtain a weak solution.

\subsection{Algorithms for Parallel Search}
Alpha-beta search is an algorithm that sequentially performs depth-first search of a game graph, and naive parallelization does not improve search efficiency very much. Many algorithms have been developed for efficient parallelization. Young Brothers Wait Concept (YBWC) \cite{feldmann1990distributed} and Lazy SMP \cite{ostensen2016completelazysmp} are popular methods for shared memory environments (e.g., a single computer). Algorithms suited for distributed memory environments (e.g., supercomputers or cloud computing) might include Asynchronous Parallel Hierarchical Iterative Deepening (APHID) \cite{brockington1998asynchronousAPHID} and ABDADA \cite{ABDADA1996}. However, distributed memory environments greatly differ due to various factors including the bandwidth and latency of interconnects between nodes, and thus the appropriate algorithms may also differ. Therefore, developers who work with current or future distributed memory environments may need to choose or develop algorithms that are appropriate for those environments.

\section{Methods}

\subsection{Use of Terms ``Ply'' and ``Move''}

In chess, there is a tradition where two sequential moves, one from white and the other from black, are called a ``move'' or ``full-move'', while an individual move is called a ``ply'' or ``half-move''. However, in this article, we refrain from using ``ply'' and always use ``move'' to denote an individual move.

\subsection{Rules of Othello}

The rules of Othello are as follows:
\begin{enumerate}
\item Black moves first, after which the players alternate.
\item If there is an empty square that satisfies the conditions for placing a stone (see below), the player to move must choose one of such empty squares and place a stone there. If there is no empty square that satisfies the condition, the player must pass their turn.
\item If a player places a stone such that there are opponent's stones in a straight line (horizontal, vertical, or diagonal) between this new stone and another of the player's stones, with no empty squares in between, then the opponent's stones in that line are flipped to become the player's stones.
\item Each player can only place a stone on the squares where placing a stone flips one or more of the opponent's stones.
\item The game ends when the board is completely filled or when there is no square on which either player can place a stone.
\item The player with more stones on the board at the end of the game wins.
\end{enumerate}
The difference in the number of stones at the end of the game is called the ``score''.

\subsection{Modification to Edax}
Existing Othello software called Edax \cite{edax2021github} was used to solve the position with 36 empty squares. Edax is based on alpha-beta search \cite{knuth1975analysis-alphabeta} and employs many techniques to improve search efficiency. While Edax is among the strongest software under typical match rules (e.g., 10 seconds per one move), its algorithm was considered suboptimal for the purpose of solving games over tens of minutes in situations where there is a large difference in scores. Therefore, the following two modifications were made:
\begin{itemize}
\item We disabled aspiration search during iterative deepening. In other words, when narrowing down alpha and beta for a solving, we modified it to take wider alpha and beta values during the shallow iterations of iterative deepening. This is because there is a tendency for the computation time to increase when the principal variation is updated. Therefore, during shallow iterations, even if a move results in a fail-high, the search should not be terminated, and instead, the best move should be sought.

\item When performing move ordering, if the results from a shallow search are found in the transposition table, we ignore them if they are relatively too shallow. This is because, especially in a solving, the search depth becomes very high. In regions where the search depth is shallow, the accuracy of move ordering based on the transposition table greatly impacts performance.
\end{itemize}

The source code of modified Edax is available on GitHub and Zenodo (see Data Availability section).

\subsection{Obtaining a set of target positions (with 50 empty squares) by optimal alpha-beta search}
\label{section:50value}

\begin{algorithm}[htb]
    \caption{$G_{50}(p,D_{50})$: Generate subset of positions with 50 empty squares as sub-problems.}
    \label{algo:gen_subset50}
    \begin{algorithmic}[1]
    \Require $p$: A position.
    \Require $D_{50}$: A dictionary where the keys are all positions with 50 empty squares, and the values are their respective predictive scores.
    \Ensure set of positions such that if all positions in it are solved and all solutions match the predictions, the initial position is consequently solved.
    \State $M \leftarrow$ a list of all legal moves on $p$
    \If{$M$ is empty}
    \State $p_{next} \leftarrow$ the position after applying pass to $p$ 
    \State $M' \leftarrow$ a list of all legal moves on $p_{next}$
    \If{$M'$ is empty}
    \State \Return $\{\}$ \Comment{Return an empty set.}
    \EndIf 
    \State \Return $G_{50}(p_{next},D_{50})$
    \EndIf 
    \If{$p$ has 50 empty squares}
    \State \Return $\{p\}$ \Comment{Return a set consisting of only $p$.}
    \EndIf
    \State $m, s \leftarrow$ the best move $m$ and corresponding score $s$ \Comment{Perform another search from $p$ to positions in $D_{50}$ to obtain $m$ and $s$.}
    \If{$s>0$}
    \State $p_{next} \leftarrow$ the position after applying $m$ to $p$ 
    \State \Return $G_{50}(p_{next},D_{50})$ \Comment{Fail-high always occurs.}
    \EndIf
    \State $l \leftarrow \{\}$ \Comment{An empty set.}
    \For{$m \in M$}
    \State $p_{next} \leftarrow$ the position after applying $m$ to $p$ 
    \State $l \leftarrow l + G_{50}(p_{next},D_{50})$ \Comment{Fail-high never occurs.}
    \EndFor
    \State \Return $l$
    \end{algorithmic}
\end{algorithm}

We developed an algorithm that requires predictive scores for all positions with 50 empty squares and returns a subset such that if all positions belonging to that subset are solved and all solutions match the predictions, the initial position is consequently solved. This is described by Algorithm \ref{algo:gen_subset50}. This algorithm is similar to the alpha-beta search with $(\alpha,\beta)=(-1,1)$ fixed. By inputting the initial position as the first argument and the dictionary of the positions and predictive scores as the second, a subset satisfying the conditions can be obtained. Notably, to keep the number of elements in the subset small, this algorithm internally performs the other alpha-beta search. As is commonly known, alpha-beta search is most efficient when the best move is searched first. Therefore, an internal alpha-beta search is conducted to find the best move. This inner search can be made more efficient through elementary memoization. As a result, even when implementing the inner search, it's possible to ensure that the computational time complexity does not increase.

\subsection{Obtaining a set of target positions (with 36 empty squares) by optimal alpha-beta search}

\begin{algorithm}[htb]
    \caption{$E(p, D')$: Estimate Game-Theoretic Value of the Given Position.}
    \label{algo:Eval}
    \begin{algorithmic}[1]
    \Require $p$: A position.
    \Require $D'$: A dictionary; key is a position and value is an estimation of its game-theoretic value.
    \Ensure An integer that is an estimation of game-theoretic value of $p$.
    \If{$p \in D'$}
    \State \Return $D'[p]$
    \EndIf 
    \State $f \leftarrow$ Edax's static evaluation function
    \State $v \leftarrow f(p)$
    \If{$|v| > 10$}
    \State \Return $v$
    \EndIf 
    \State $\alpha \leftarrow \min(-3,v)$
    \State $\beta \leftarrow \max(3,v)$
    \State \Return the value determined by an alpha-beta search from $p$ to a depth of 2, using $f$ as the evaluation function for leaf nodes and $\alpha,\beta$ as initial alpha and beta.
    \end{algorithmic}
\end{algorithm}

\begin{algorithm}[htb]
    \caption{$G_{36}^{third}(p,D,k,\alpha,\beta)$: Calculate an upper or lower bound of Game-Theoretic Value of the Given Position.}
    \label{algo:36third}
    \begin{algorithmic}[1]
    \Require $p$: A position.
    \Require $D$: A dictionary; key is a position and value is the exact upper and lower bounds of its game-theoretic value.
    \Require $k$: A Boolean that indicates the kind of search, i.e., upper or lower.
    \Require $\alpha,\beta$: Parameters for alpha-beta search.
    \Ensure An integer that is an upper (or lower; determined by $k$) bound of the game-theoretic value of $p$.
    \State $M \leftarrow$ a list of all legal moves on $p$.
    \If{$M$ is empty}
    \State $p_{next} \leftarrow$ the position after applying pass to $p$ 
    \State $M' \leftarrow$ a list of all legal moves on $p_{next}$
    \If{$M'$ is empty}
    \State \Return the final value of $p$. \Comment{The game is ended.}
    \EndIf 
    \State \Return $-G_{36}^{third}(p_{next},D,\neg k,-\beta,-\alpha)$
    \EndIf
    \If{the number of empty squares is 36}
    \If{$p \in D$}
    \If{k}
    \State \Return $D[p]$.upperbound \Comment{Return an upper bound of game-theoretic value of $p$.}
    \EndIf 
    \State \Return $D[p]$.lowerbound \Comment{Return an lower bound of game-theoretic value of $p$.}
    \EndIf 
    \If{there is a move that causes wipe-out}
    \State \Return 64
    \EndIf 
    \State \Return 64
    \EndIf 
    \State Sort $M$ in descending order of how promising the result is using a deterministic method.
    \State $v \leftarrow$ an empty list.
    \For{$m \in M$}
    \State $p_{next} \leftarrow$ the position after applying $m$ to $p$ 
    \State $v \leftarrow v + -G_{36}^{third}(p_{next},D,\neg k,-\beta,-\alpha)$ \Comment{Nega-max search. Nega-scout can be used.}
    \If{$\max(v) \geq \beta$}
    \State \Return $\max(v)$ \Comment{Fail-high occurred.}
    \EndIf 
    \EndFor
    \State \Return $\max(v)$
    \end{algorithmic}
\end{algorithm}

\begin{algorithm}[htb]
    \caption{$G_{36}^{second}(p,D,D',\alpha,\beta)$: Calculate an estimation of Game-Theoretic Value of the Given Position.}
    \label{algo:36second}
    \begin{algorithmic}[1]
    \Require $p$: A position.
    \Require $D$: A dictionary; key is a position and value is the exact upper and lower bounds of its game-theoretic value.
    \Require $D'$: A dictionary; key is a position and value is an estimation of its game-theoretic value.
    \Require $\alpha,\beta$: Parameters for alpha-beta search.
    \Ensure An integer that is an estimation of game-theoretic value of $p$.
    \State $M \leftarrow$ a list of all legal moves on $p$.
    \If{$M$ is empty}
    \State $p_{next} \leftarrow$ the position after applying pass to $p$ 
    \State $M' \leftarrow$ a list of all legal moves on $p_{next}$
    \If{$M'$ is empty}
    \State \Return the final value of $p$. \Comment{The game is ended.}
    \EndIf 
    \State \Return $-G_{36}^{second}(p_{next},D,D',-\beta,-\alpha)$
    \EndIf
    \If{the number of empty squares is 36}
    \If{there is a move that causes wipe-out}
    \State \Return 64
    \EndIf 
    \State $e \leftarrow E(p,D')$ \Comment{Call algorithm \ref{algo:Eval}.}
    \If{$p \in D$}
    \If{$D[p]$.lowerbound $= D[p]$.upperbound}
    \State \Return $D[p]$.upperbound \Comment{Return an exact game-theoretic value of $p$.}
    \ElsIf{$\min(\beta,e) \leq D[p]$.lowerbound}
    \State \Return $D[p]$.lowerbound \Comment{Return a lower bound of game-theoretic value of $p$.}
    \ElsIf{$D[p]$.upperbound $\leq \max(\alpha,e)$}
    \State \Return $D[p]$.upperbound \Comment{Return a upper bound of game-theoretic value of $p$.}
    \EndIf 
    \State \Return $e$ \Comment{Return an estimation of game-theoretic value of $p$.}
    \EndIf 
    \State \Return 64
    \EndIf 
    \State Sort $M$ in descending order of how promising by a deterministic method.
    \State $v \leftarrow$ an empty list.
    \For{$m \in M$}
    \State $p_{next} \leftarrow$ the position after applying $m$ to $p$ 
    \State $v \leftarrow v + -G_{36}^{second}(p_{next},D,D',-\beta,-\alpha)$ \Comment{Nega-max search. Nega-scout can be used.}
    \If{$\max(v) \geq \beta$}
    \State \Return $\max(v)$ \Comment{Fail-high occurred.}
    \EndIf 
    \EndFor
    \State \Return $\max(v)$
    \end{algorithmic}
\end{algorithm}

\begin{algorithm}[htb]
    \caption{$G_{36}^{first}(p,D,D',A,\alpha,\beta)$: Traverse Game-Graph and Obtain Positions to Solve.}
    \label{algo:36first}
    \begin{algorithmic}[1]
    \Require $p,D,D',\alpha,\beta$: The same parameters as in Algorithm \ref{algo:36second}.
    \Require $A$: A dictionary in which key is a position with 36 empty square and value is a tuple; each tuple consists of an estimated game-theoretic value and two integers $\alpha$ and $\beta$. To solve $p$, we should solve all positions in the key under the $\alpha$ and $\beta$ in the value. If all estimations are correct, then $p$ is solved.
    \Ensure An integer that is an estimation of game-theoretic value of $p$.
    \Ensure The updated $A$.
    \State $M \leftarrow$ a list of all legal moves on $p$.
    \If{$M$ is empty}
    \State $p_{next} \leftarrow$ the position after applying pass to $p$ 
    \State $M' \leftarrow$ a list of all legal moves on $p_{next}$
    \If{$M'$ is empty}
    \State \Return the final value of $p$, and $A$. \Comment{The game is ended.}
    \EndIf 
    \State \Return $-G_{36}^{first}(p,D,D',A,-\beta,-\alpha)$
    \EndIf
    \If{the number of empty squares is 36}
    \If{there is a move that causes wipe-out}
    \State \Return 64,$A$
    \EndIf 
    \State $e \leftarrow G_{36}^{second}(p,D,D',\alpha,\beta)$
    \If{$p \notin D$}
    \State $A[p] \leftarrow (\alpha,\beta,e)$
    \Else
    \State $A[p] \leftarrow $ ($\min(\alpha,A[p].\alpha),\max(\beta,A[p].\beta),e$)
    \EndIf 
    \State \Return $e,A$ \Comment{Return an estimation of game-theoretic value of $p$.}
    \EndIf 
    \State $v \leftarrow$ an empty list.
    \State $S \leftarrow$ an empty dictionary.
    \For{$m \in M$}
    \State $p_{next} \leftarrow$ the position after applying $m$ to $p$ 
    \State $e_{lower} \leftarrow -G_{36}^{third}(p_{next},D,True,-64,64)$\Comment{True means upper-mode.}
    \State $e_{upper} \leftarrow -G_{36}^{third}(p_{next},D,False,-64,64)$\Comment{False means lower-mode.}
    \If{$\beta\leq e_{lower}$}
    \State \Return $e_{lower}, A$ \Comment{Fail-high occurred.}
    \ElsIf{$e_{upper}\leq \alpha$}
    \State $v \leftarrow v + e_{upper}$
    \State remove $m$ from $M$.
    \State continue.
    \ElsIf{$\alpha < e_{lower} = e_{upper} < \beta$}
    \State $v \leftarrow v + e_{upper}$
    \State $\alpha \leftarrow e_{upper}$
    \State remove $m$ from $M$.
    \State continue.
    \EndIf
    \State $e \leftarrow -G_{36}^{second}(p_{next},D,D',-64,64)$
    \State $S[m] \leftarrow e$ \Comment{We can add $S[m]$ to some auxiliary heuristic factors to improve the following move ordering.}
    \EndFor
    \If{$M$ is empty}
    \State \Return $\max(v), A$
    \EndIf
    \State Sort $M$ in descending order of the values in $S$.
    \For{$m \in M$}
    \State Perform the same alpha-beta search as in Algorithms \ref{algo:36third} and \ref{algo:36second}.
    \EndFor
    \State \Return $\max(v), A$
    \end{algorithmic}
\end{algorithm}

We implemented Algorithm \ref{algo:36first}, which requires a position with 50 empty squares and data about position(s) with 36 empty squares and outputs a set of position(s) with 36 empty squares and a corresponding result hypothesis. This algorithm can process known search outcomes for positions with 36 empty squares, and output position(s) with 36 empty squares and corresponding estimated game-theoretic value; if we can confirm that all outputted estimations are correct, then we can prove the game-theoretic value of the input position. Importantly, it can differentiate between positions that we have obtained game-theoretic value for and those that we have only estimated the value of.

Algorithms \ref{algo:Eval}, \ref{algo:36third}, and \ref{algo:36second} are auxiliary algorithms for Algorithm \ref{algo:36first}. These three algorithms can be made more efficient through elementary memoization, which is omitted from the pseudo-codes. The source code is available at GitHub (see Data Availability section).

We solved the positions with 36 empty squares, which were obtained from Algorithm \ref{algo:36first}, using a computer cluster and Edax software. If the predicted value of the results was less than 30, we read with 4 cores; otherwise, we did so with 1 core. In the initial phase of this computation, we did not have confidence that the game-theoretical value of the initial position would be a draw, so we set the alpha-beta window to $[-3, +3]$. Subsequently, when reading with 4 cores, we changed it to read with a $[-1, +1]$ window. When reading with 1 core, we always kept the window at $[-3, +3]$. The advantage of reading with 1 core is avoiding efficiency degradation due to parallel searching, and because the number of search positions becomes deterministic, it ensures reproducibility. The continued benefit of keeping the window at $[-3, +3]$ is that if the game-theoretical value of the initial position turned out to be -2 rather than a draw, there would be no need for re-computation, and there is no need to change the command-line options of Edax, meaning we do not have to save it for each problem. The downside is a possible slight increase in the number of search positions, but for positions with a significant difference, it is believed to have minimal impact compared to when the window is set to $[-1, +1]$.

\begin{algorithm}[htb]
    \caption{$Q(p,\alpha,\beta)$: Calculate the Game-Theoretic Value of the Given Position.}
    \label{algo:50loop}
    \begin{algorithmic}[1]
    \Require $p$: A position.
    \Require $\alpha,\beta$: Parameters for alpha-beta search.
    \Ensure An integer that is the game-theoretic value of $p$.
    \Ensure A dictionary; key is a position and value is the exact upper and lower bounds of its game-theoretic value. One can prove $p$ only from the information in this dictionary.
    \State $D \leftarrow$ an empty dictionary.
    \State $D' \leftarrow$ an empty dictionary.
    \While{$True$}
    \State $A \leftarrow$ an empty dictionary.
    \State $v,A \leftarrow G_{36}^{first}(p,D.D',A,\alpha,\beta)$
    \If{$A$ is empty}
    \State \Return $v,D$
    \EndIf
    \State $R \leftarrow$ The game-theoretic values (or its estimation) of all positions in $A$. \Comment{Edax is available.}
    \State $D \leftarrow D+$ The exact game-theoretic values in $R$.
    \State $D' \leftarrow D'+$ The estimations in $R$.
    \EndWhile
    \end{algorithmic}
\end{algorithm}

The game-theoretic values of the output positions can sometimes deviate from the estimations of the above algorithm. In such cases, by adding the results to $D$ and rerunning the Algorithm \ref{algo:36first}, more positions to be solved can be identified. Once Algorithm \ref{algo:36first} no longer outputs any position, it can be said that the proof of the input position with 50 empty squares has been established. This procedure can be described as Algorithm \ref{algo:50loop}.

\subsection{Constructing a program that never loses}

To satisfy the Weakly solved condition, we implemented a Python script that acts as a perfect (i.e., never loses) player by referring to the results. The script refers to the result table while there are more than 36 empty squares, and after that, it delegates to Edax to play perfectly.

\subsection{Materials}

To solve positions with 36 empty squares, we used the CPUs of MN-J, a supercomputer owned by Preferred Networks Inc. MN-J refers collectively to multiple supercomputers: MN-2A, MN-2B, and MN-3, all of which appear to users as a single Kubernetes cluster. This supercomputer is equipped with several types of CPUs (Intel Xeon 6254, AMD EPYC 7713, Intel Xeon 8380, and Intel Xeon 8260M), and all CPUs feature main memory with Error Checking and Correction (ECC).

\clearpage

\section{Results}


First of all, we enumerated and shortly evaluated all positions with 50 empty squares. We only enumerated positions with at least one legal move and considered symmetrical positions to be identical. As a result, 2,958,551 positions were enumerated. We evaluated all of them by Edax for 10 seconds using a single CPU core. For positions that resulted in values close to a draw from the 10-second evaluations, we conducted more extended evaluations.

Next, we selected 2,587 positions out of the 2,958,551 positions and formulated hypotheses regarding their game-theoretic values. We chose them such that if all these hypotheses were proven correct, it would prove that the initial position results in a draw. Although there are numerous ways to select subsets that would prove that the initial position results in a draw, we used Algorithm \ref{algo:gen_subset50} to obtain a small subset. For the evaluation values, we used the values obtained from the previously mentioned evaluations. In cases where the values were the same, we prioritized positions that appear frequently in the WTHOR database\cite{wthor-french-database} of Othello games published by the French Othello Federation. We used a dataset including 61,549 game records played between 2001 and 2020. As we will describe in detail later, it was proven that all these 2,587 hypotheses were correct.

\begin{figure}[htb]
\centering
\includegraphics[width=\linewidth]{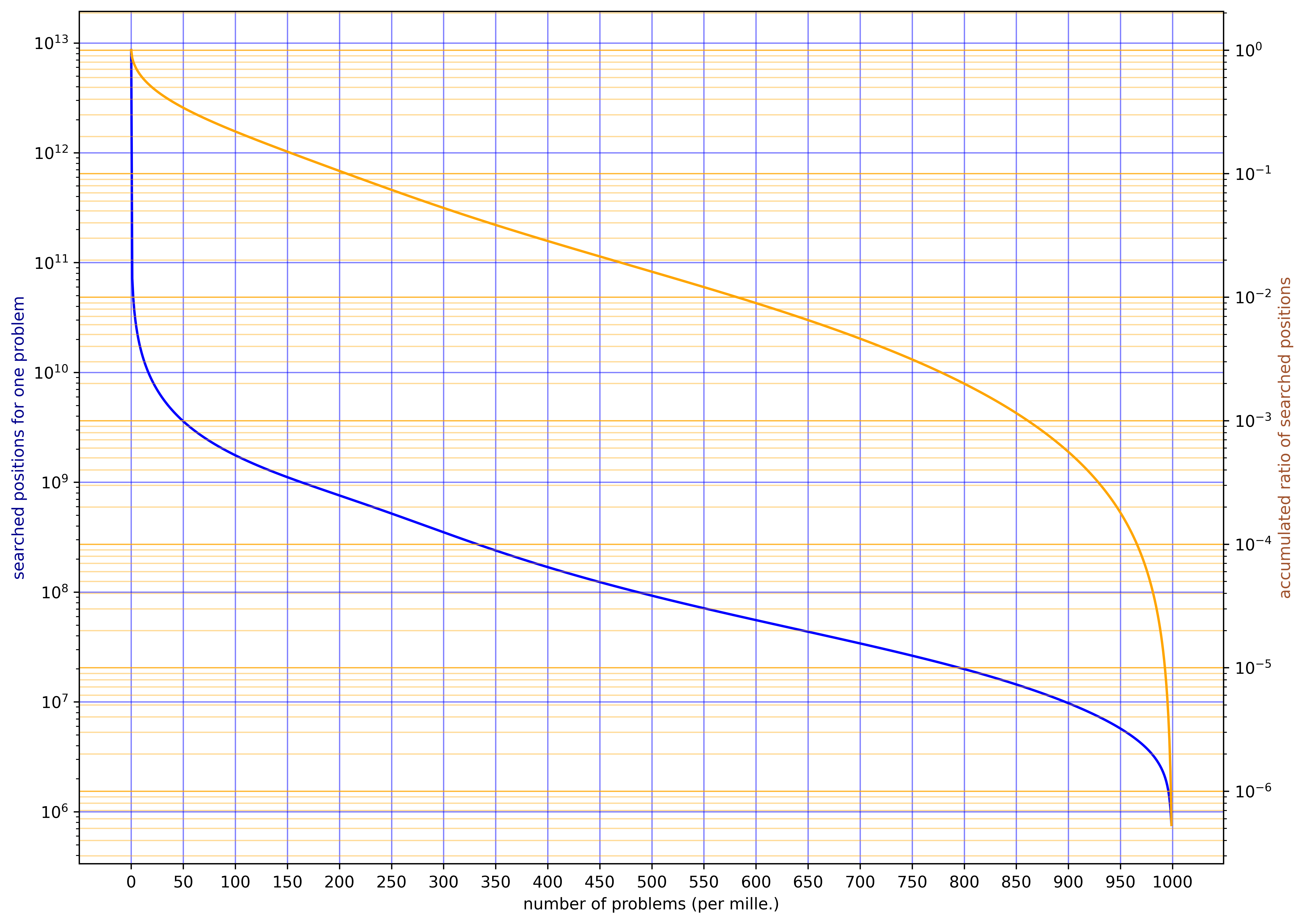}
\caption{
Positions with 36 empty squares that were solved to prove the initial position were sorted in descending order by the number of search phases reported by Edax, and the cumulative number (orange) and number of search phases for a single particular question (blue) were plotted for each $1/1000$.
}
\label{fig:fig_1000_accumulateion}
\end{figure}

\begin{table}
 \caption{Problem size and search capability}
  \centering
  \begin{tabular}{lll}
    \toprule
    Factor     & Description     & Number \\
    \midrule
    Problems   & Number of solved problems to wealky solve Othello  & $\sim 1.5 \times 10^{9}$\\
    CPU        & Searching capability using Edax (positions/GHz/core/sec) & $\sim 1.2 \times 10^{7}$\\
    Positions  & Number of searched positions (reported by Edax) & $\sim 1.5 \times 10^{18}$\\
    \bottomrule
  \end{tabular}
  \label{tab:table}
\end{table}

As a result, the number of positions with 36 empty squares needed to solve the initial position amounted to $1,505,367,525$, with the total search positions reported by Edax for all these positions reaching $1,526,001,455,595,489,506$ (Figure \ref{fig:fig_1000_accumulateion}, Table \ref{tab:table}). Due to having the alpha-beta window set to $[-3, +3]$ for some borderline positions, there seems to be room for reduction in this number. As a null-window search is available for verification, the number of necessary search positions could potentially be even lower.

\begin{figure}[htb]
\centering
\includegraphics[width=\linewidth]{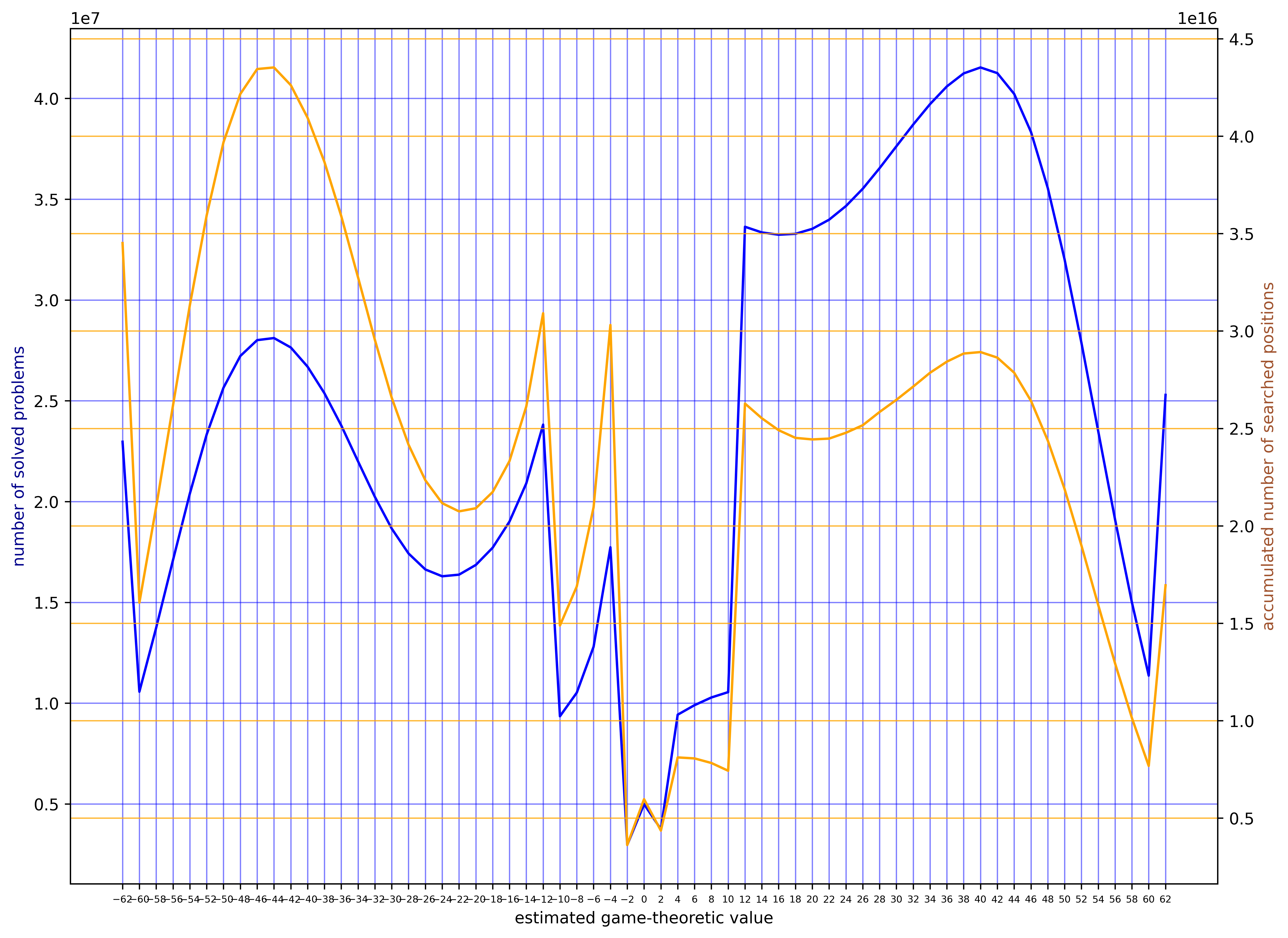}
\caption{
Positions with 36 empty squares that were solved to prove the initial position were classified according to the value of Algorithm \ref{algo:Eval} (horizontal axis) and the sum of the numbers of searched positions reported by Edax was calculated for each (vertical axis).
}
\label{fig:fig_estimated_value}
\end{figure}

\begin{figure}[htb]
\centering
\includegraphics[width=0.8\linewidth]{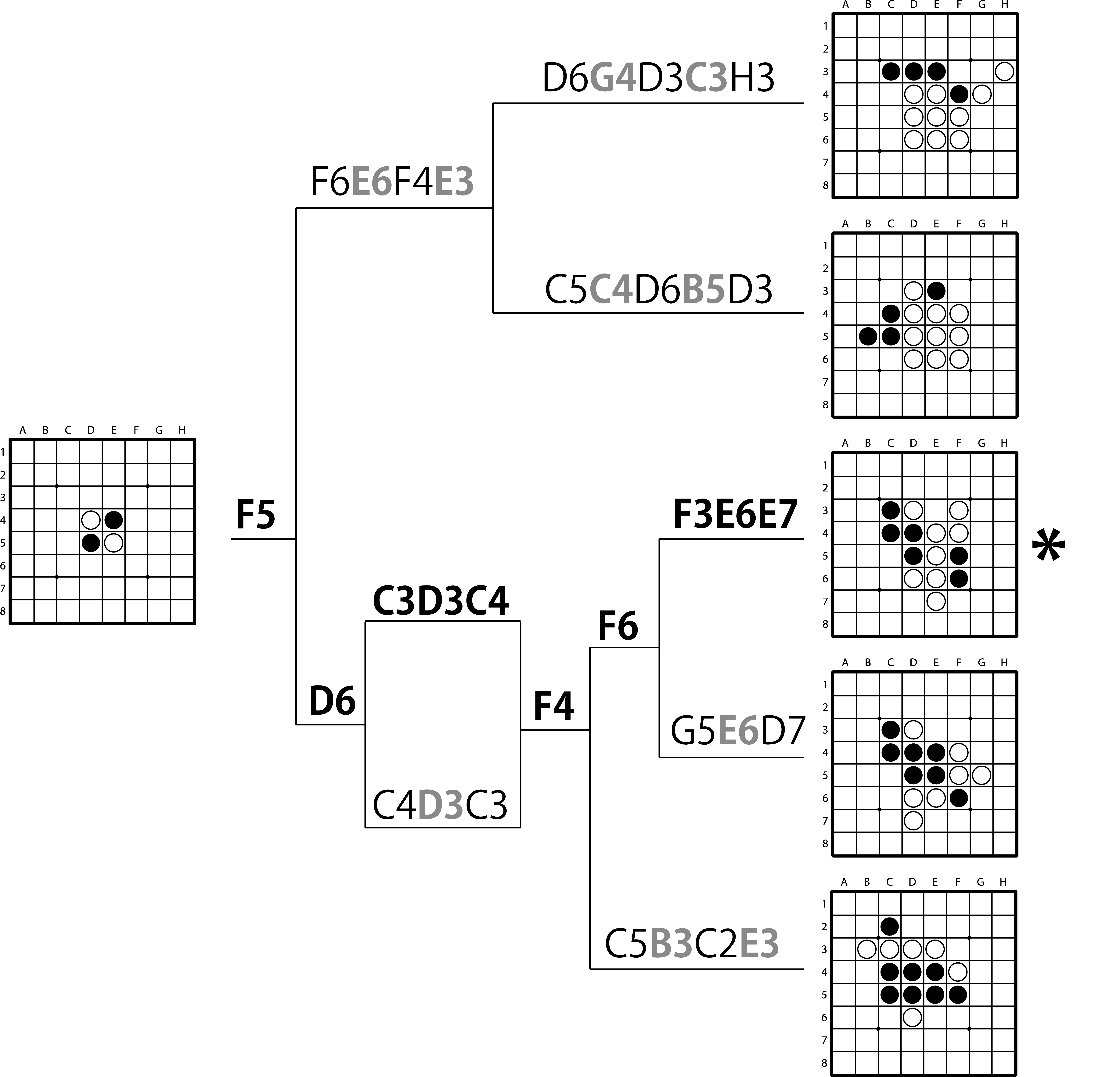}
\caption{A graphical representation of results about opening of Othello. The bold black moves show the optimal game record. Our perfect player always chooses the bold (black or gray) moves in the corresponding position. Right five positions are proved that all those game-theoretic values are draws. Center one with asterisk is the progress of Figure \ref{fig:fig_game_record}.
}
\label{fig:fig_opening_graph}
\end{figure}

The results of the opening are illustrated in Figure \ref{fig:fig_opening_graph}. Our perfect player never voluntarily deviates from the optimal game record (shown as bold black moves). If the opponent chooses a move not shown in this figure, we proved that our player will always win. If the opponent chooses one of the non-bold moves, we proved that our player draws or wins by choosing moves shown as bold gray moves.

\section{Discussion and Conclusions}

We conclude that our study has weakly solved Othello, although we recognize that our achievement is just above the criteria for weakly solving. For certain borderline positions with 36 empty squares, Edax requires a large amount of computation to determine the game-theoretic value and corresponding move. However, given the continuing advances in personal computers, it is reasonable to conclude that our approach requires only reasonable computational resources. By providing an additional ``opening'' book for these positions with 35 or fewer empty squares, we could further reduce the computational demand. However, to expedite our announcement, we opted against computing additional books in this study. Nonetheless, there may be interest among Othello enthusiasts for software that can determine the best move using fewer computational resources.

As Figure \ref{fig:fig_estimated_value} indicates, many of our calculations to weakly solve Othello were devoted to positions where, according to the estimation, there is a clear advantage in terms of winning or losing. This indicates that one cannot claim a pseudo-solution by not proving positions whose estimated game-theoretic value exceeds any threshold. 

As Figure \ref{fig:fig_estimated_value} implies, for positions that were expected to have a significant difference in scores, there were some in which a solving by Edax took a large amount of time. It is possible that the systematic and significant errors in Edax's function to estimate the game-theoretic value from a position (called the static evaluation function), especially for positions unlikely to appear in actual games, are the reason for this. Regarding this issue, while it can be addressed by preparing an additional opening book, there is also potential for retraining or improving the design of the static evaluation function.

We recognize that some readers may be skeptical about the validity of computational proofs. Naturally, computational errors due to CPU or memory faults cannot be entirely ruled out. However, as the vast majority of calculations were executed on a computer cluster with ECC memory, we believe the results to be nearly indisputable. Moreover, even if a computational error were present, the chance of overturning our conclusion of a final draw is extremely low. If any errors are detected, they can be easily recalculated using the publicly released software.

To the best of our knowledge, no category in between weakly and strongly solving has been proposed. We considered strongly solving Othello is intractable and aimed for a weak solution. We have created software that will always achieve a draw or win to achieve the criteria for weakly solving. If the opponent makes a blunder, however, we do not guarantee that the software capitalizes on it. 

Although strongly solving the game may be intractable, developing software that consistently makes the best move represents a challenge that lies between weak and strong solving, and is likely to attract widespread interest. Therefore, we would propose to call this intermediate category "semi-strong solving". This study does not achieve semi-strong solving of Othello; this remains as future work.

Considering the game's popularity and estimated size of the search space, we speculate that chess might be the next weakly solved grand challenge. However, because the search space of chess is very large, not only improvements in computational power but also theoretical breakthroughs might be necessary. We hope that this study will inspire readers and contribute to significant advancements in future computer science.

\section*{Acknowledgments}

The author gratefully thanks members of Preferred Networks Inc., including Dr. Kenta Oono, Dr. Kohei Hayashi, Dr. Masanori Koyama, and Dr. Shin-ichi Maeda, for their useful discussions and encouragement.

Parts of this study, especially the majority of the calculations, were conducted during the author's work time at Preferred Networks Inc. This was made possible by the company's 20-percent rule, which allows employees to dedicate 20 percent of their work time to pursue their own ideas and projects. The author gratefully thanks the company for having the rule.

\section*{Additional Information and Declarations}

\subsection*{Competing Interests}

The author declares that there are no competing interests.

\subsection*{Author Contributions}

Hiroki Takizawa conceived and designed the research, implemented and performed the computational experiments, analyzed the data, prepared figures and tables, authored drafts of the paper, and approved the final draft.

\subsection*{Data Availability}

The source code of modified Edax is available at GitHub (https://github.com/eukaryo/edax-reversi-AVX-v446mod2), and Zenodo (https://doi.org/10.5281/zenodo.10030906).

The raw outputs of analyses are available at figshare (https://doi.org/10.6084/m9.figshare.24420619).

The source code for analyses are available at GitHub (https://github.com/eukaryo/reversi-scripts).

\subsection*{Funding}

The author did not receive any academic funding for this study.

\bibliographystyle{unsrt}  
\bibliography{references}

\begin{thebibliography}{10}

\bibitem{babbage1864passages}
Charles Babbage.
\newblock {\em Passages from the Life of a Philosopher}.
\newblock London: Longman, 1864.

\bibitem{shannon1950xxii}
Claude~E Shannon.
\newblock Xxii. programming a computer for playing chess.
\newblock {\em The London, Edinburgh, and Dublin Philosophical Magazine and
  Journal of Science}, 41(314):256--275, 1950.

\bibitem{campbell2002deepblue}
Murray Campbell, A.~Joseph Hoane, and Feng-hsiung Hsu.
\newblock Deep blue.
\newblock {\em Artif. Intell.}, 134(1–2):57–83, January 2002.

\bibitem{silver2016mastering-alphago}
David Silver, Aja Huang, Christopher~J. Maddison, Arthur Guez, Laurent Sifre,
  George van~den Driessche, Julian Schrittwieser, Ioannis Antonoglou, Veda
  Panneershelvam, Marc Lanctot, Sander Dieleman, Dominik Grewe, John Nham, Nal
  Kalchbrenner, Ilya Sutskever, Timothy Lillicrap, Madeleine Leach, Koray
  Kavukcuoglu, Thore Graepel, and Demis Hassabis.
\newblock Mastering the game of go with deep neural networks and tree search.
\newblock {\em nature}, 529(7587):484--489, 2016.

\bibitem{kaneko2019shogi}
Tomoyuki Kaneko and Takenobu Takizawa.
\newblock Computer shogi tournaments and techniques.
\newblock {\em IEEE Transactions on Games}, 11(3):267--274, 2019.

\bibitem{alphazeropaper}
David Silver, Thomas Hubert, Julian Schrittwieser, Ioannis Antonoglou, Matthew
  Lai, Arthur Guez, Marc Lanctot, Laurent Sifre, Dharshan Kumaran, Thore
  Graepel, Timothy Lillicrap, Karen Simonyan, and Demis Hassabis.
\newblock A general reinforcement learning algorithm that masters chess, shogi,
  and go through self-play.
\newblock {\em Science}, 362(6419):1140--1144, 2018.

\bibitem{buro1997othello}
Michael Buro.
\newblock The othello match of the year: Takeshi murakami vs. logistello.
\newblock {\em ICGA Journal}, 20(3):189--193, 1997.

\bibitem{allis1994searching-thesis}
L.~V. Allis.
\newblock {\em Searching for Solutions in Games and Artificial Intelligence}.
\newblock PhD thesis, Department of Computer Science, University of Limburg,
  1994.

\bibitem{takizawa2023strongly}
Hiroki Takizawa.
\newblock Strongly solved ostle: calculating a strong solution helps compose
  high-quality puzzles for recent games.
\newblock {\em PeerJ Computer Science}, 9:e1560, 2023.

\bibitem{schaeffer2007checkers}
Jonathan Schaeffer, Neil Burch, Yngvi Bj{\"o}rnsson, Akihiro Kishimoto, Martin
  M{\"u}ller, Robert Lake, Paul Lu, and Steve Sutphen.
\newblock Checkers is solved.
\newblock {\em science}, 317(5844):1518--1522, 2007.

\bibitem{gasser1996solving-nine-mens-morris}
Ralph Gasser.
\newblock Solving nine men's morris.
\newblock {\em Computational Intelligence}, 12(1):24--41, 1996.

\bibitem{romein2003solving-awari}
John~W Romein and Henri~E Bal.
\newblock Solving awari with parallel retrograde analysis.
\newblock {\em Computer}, 36(10):26--33, 2003.

\bibitem{knuth1975analysis-alphabeta}
Donald~E Knuth and Ronald~W Moore.
\newblock An analysis of alpha-beta pruning.
\newblock {\em Artificial intelligence}, 6(4):293--326, 1975.

\bibitem{thompson1986retrograde}
Ken Thompson.
\newblock Retrograde analysis of certain endgames.
\newblock {\em J. Int. Comput. Games Assoc.}, 9(3):131--139, 1986.

\bibitem{nagai2002df-pn}
Ayumu Nagai.
\newblock {\em Df-pn algorithm for searching AND/OR trees and its
  applications}.
\newblock PhD thesis, Department of Information Science, University of Tokyo,
  2002.

\bibitem{kishimoto2012game}
Akihiro Kishimoto, Mark~HM Winands, Martin M{\"u}ller, and Jahn-Takeshi Saito.
\newblock Game-tree search using proof numbers: The first twenty years.
\newblock {\em Icga Journal}, 35(3):131--156, 2012.

\bibitem{ALLIS1994pnsearch}
L.Victor Allis, Maarten {van der Meulen}, and H.Jaap {van den Herik}.
\newblock Proof-number search.
\newblock {\em Artificial Intelligence}, 66(1):91--124, 1994.

\bibitem{feldmann1990distributed}
Rainer Feldmann, Burkhard Monien, Peter Mysliwietz, and Oliver Vornberger.
\newblock Distributed game tree search.
\newblock {\em Parallel Algorithms for Machine Intelligence and Vision}, pages
  66--101, 1990.

\bibitem{ostensen2016completelazysmp}
Emil~Fredrik {\O}stensen.
\newblock A complete chess engine parallelized using lazy smp.
\newblock Master's thesis, University of Oslo, 2016.

\bibitem{brockington1998asynchronousAPHID}
Mark~Gordon Brockington.
\newblock {\em Asynchronous Parallel Garne-Tree Search}.
\newblock PhD thesis, University of Alberta, 1998.

\bibitem{ABDADA1996}
Jean-Christophe Weill.
\newblock The abdada distributed minimax search algorithm.
\newblock In {\em Proceedings of the 1996 ACM 24th Annual Conference on
  Computer Science}, CSC '96, page 131–138, New York, NY, USA, 1996.
  Association for Computing Machinery.

\bibitem{edax2021github}
Richard Delorme.
\newblock edax-reversi.
\newblock https://github.com/abulmo/edax-reversi, 2021.
\newblock Last retrieved 2023-07-07.

\bibitem{wthor-french-database}
French~Othello Federation.
\newblock La base wthor.
\newblock https://www.ffothello.org/informatique/la-base-wthor/, 2021.
\newblock Last retrieved 2023-10-13.

\end{thebibliography}

\end{document}